\documentclass{article}

\usepackage[preprint]{sty/corl_2026} 

\usepackage{graphicx}
\usepackage{bm}
\usepackage{tikz}
\usetikzlibrary{calc,math}
\usepackage{algorithm}
\usepackage[noEnd=true,indLines=true,commentColor=black]{sty/algpseudocodex}
\usepackage{cleveref}
\usepackage{siunitx}
\usepackage{booktabs}
\usepackage{wrapfig}
\usepackage{sty/custom}
\usepackage{subcaption}
\usepackage{svg}

\title{Learning Coordinated Visuomotor Box-Pushing\\from Solo Demonstrations}

\author{
  Ryusei Matsumoto$^{1,2}$, Mei Kinjo$^{1,3}$, Yoko Sasaki$^1$, Keisuke Okumura$^1$\\
  $^1$National Institute of Advanced Industrial Science and Technology (AIST), Japan\\  
  $^2$Institute of Science Tokyo, Japan\\
  $^3$The University of Tokyo, Japan\\
  \texttt{\{matsumoto.ryusei,y-sasaki,okumura.k\}}@aist.go.jp
}
\begin{document}
\maketitle
\begin{abstract}
Multi-robot imitation learning, particularly in settings where visuomotor policies are deployed in a communication-free, onboard decentralised style, represents an attractive paradigm.
However, its realisation remains insufficiently understood, largely due to the difficulty of collecting collective demonstrations, since a single operator cannot control many robots simultaneously.
Meanwhile, unlike coupled collaborative manipulation, many \emph{coordinated} tasks achieve system-wide efficiency primarily through minimising inter-robot interference.
This structure motivates us to study whether data collected by a teleoperated single-robot can be leveraged for large-scale \emph{coordinated box-pushing} as a testbed.
We systematically investigate dataset creation strategies and lightweight policy architectures.
In particular, experiments with up to 40 robots highlight the difficulty of acquiring effective coordination solely through passive observation of other operating robots, revealing a concrete bottleneck for multi-robot research.
\end{abstract}
{
\begin{figure}[th!]
  \centering
  \includegraphics[width=1.0\linewidth]{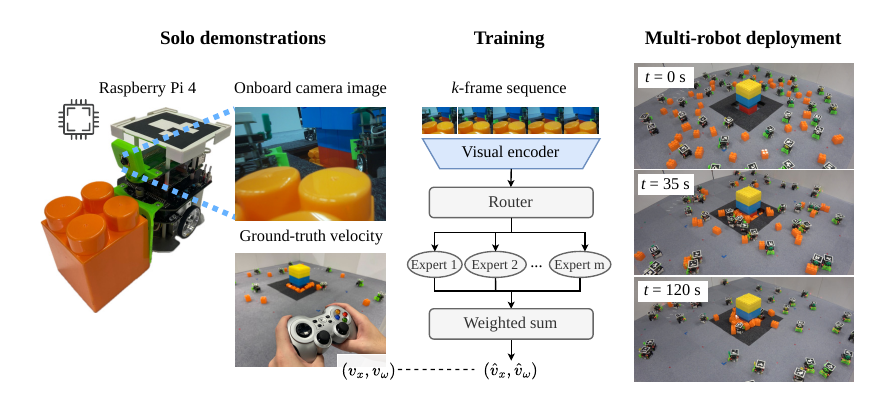}
  \caption{
    We learn coordinated box-pushing policies from solo demonstrations (\emph{left}).
    A lightweight mixture-of-experts architecture captures diverse skills under onboard resource constraints (\emph{middle}).
    The resulting policy enables a multi-robot team to clean up scattered objects (\emph{right}).
    The video is available at \url{https://youtu.be/J9KfPI-hJX4}.
  }
  \label{fig:overview}
\end{figure}
}

\section{Introduction}
Robotic learning from human demonstrations has become a powerful paradigm for acquiring capable visuomotor policies, ranging from autonomous driving to dexterous manipulation~\cite{bojarski2016end,chi2025diffusion,black2026pi_0}.
However, extending imitation learning to the multi-robot domain remains substantially more challenging.
This is because, in addition to multi-robot-specific challenges such as task decomposition~\cite{gerkey2004formal} and interaction modelling~\cite{gielis2022critical}, \emph{learning coordinated multi-robot policies suffers from the difficulty of collecting high-quality demonstration data for collective motions.}

More concretely, a closer look at modern robotic learning systems reveals that \emph{teleoperation} has become the dominant paradigm for collecting high-quality robot demonstrations~\cite{fu2024mobile}.
Although a human operator may be able to control a small number of robots simultaneously, obtaining demonstrations for large-scale collective behaviours rapidly becomes impractical due to the increasing cognitive and operational burden.
One possible solution is to employ multiple expert operators to jointly generate demonstrations; however, such an approach scales poorly as the number of robots increases.
Consequently, scalable multi-robot imitation learning has remained underdeveloped.

While a few recent studies have explored multi-robot imitation learning~\cite{mattson2025r2bc,ji2025enabling}, they typically rely on external infrastructure such as motion capture systems for globally consistent state estimation, limiting real-world deployment.
In contrast, modern single-robot systems increasingly achieve robust operation using only onboard sensing and computation~\cite{kaufmann2023champion,hoeller2024anymal}.
In a similar spirit, we argue that multi-robot coordination should likewise be achievable using compact onboard visuomotor policies that directly map pixels to actions under resource constraints.
Communication-free and fully decentralised approaches are further attractive for scalable deployment, avoiding bottlenecks such as latency, packet loss, and communication infrastructure dependencies.

These observations motivate the development of \textbf{onboard, communication-free visuomotor policies for scalable multi-robot coordination learned from \emph{solo demonstrations}}.
We use ``solo demonstrations'' to mean demonstrations in which only a single robot is directly controlled by a human and only that robot’s observation-action pairs are recorded, even when other robots are present as static or autonomous context.
Our central question is then how such policies can be effectively attained.
Focusing on a \textbf{coordinated box-pushing} task in which multiple mobile robots collect scattered objects into a designated region to clean up a workspace (\cref{fig:overview}), \textbf{this study aims to provide empirical insights into the aforementioned challenge using large-scale deployments.}

We specifically chose this task because it goes beyond mere multi-robot navigation while remaining closely connected to practical multi-robot applications such as collective construction~\cite{petersen2019review}.
Although prior works have explored vision-based multi-robot tasks such as collision avoidance~\cite{zhang2025learning} and flocking~\cite{hu2021vgai,mezey2025purely}, our setting differs substantially in both task complexity with physical interaction and system scale, up to 40 robots.
Through this setup, our contribution is a systematic real-robot study of what kinds of solo demonstrations actually induce effective coordination, revealing a nuanced picture.
Surprisingly, data collected with static peer robots provides a strong and data-efficient signal, outperforming data collected among actively moving policy-controlled robots.
However, throughput does not scale linearly with robot count, suggesting that dynamic-agent data collection remains an unresolved bottleneck.
These findings highlight scalable multi-robot learning as a worthwhile open challenge.

\section{Problem Setting}
\label{sec:task}
The task setup is illustrated in \cref{fig:overview}.
A team of differential-drive robots~\cite{yamamoto2025100mouse} collects boxes (28 in the experiments) scattered throughout the white-walled workspace and transports them to a target bin located at the centre.
The bin is marked by a tower with coloured blocks for visual recognition, and the robots dispose of the collected boxes by dropping them into a hole at the base of the bin.
The workspace layout is available in the appendix.
Each robot operates solely on a Raspberry Pi running ROS~2, without GPU acceleration, and is equipped with a front-facing camera that provides ego-centric observations with a field of view of \SI{120}{\degree}.
At each control step (\SI{5}{\hertz}), robots receive RGB image observations and output velocity commands.
Crucially, the robots operate in a communication-free, onboard decentralised manner, without relying on external infrastructure such as localisation systems.
The markers mounted on top of the robots are used solely for measuring task completion progress.

We refer to this task as ``coordinated'' box pushing to deliberately distinguish it from ``collaborative'' counterparts (e.g.,~\cite{nachum2020multi,tang2024collaborative,feng2025learning}), which require tightly coupled problem solving that cannot be accomplished by a single robot alone.
Coordinated tasks can be completed by a single robot, while deploying multiple robots can improve task throughput through concurrent operation~\cite{prorok2021beyond}.
Yet, inter-robot interference can significantly degrade performance.
For example, in our setup, robots may collide head-on and become stuck, steal boxes being transported by other robots, or even push other robots into the disposal hole; all of which were observed in practice.
In particular, the densely populated workspace often leads to congestion, making such coordination failures ubiquitous.
This distinction between coordinated and collaborative settings is fundamental to this study, as it enables us to train multi-robot policies from solo demonstrations.

\section{Related Work}
Box-pushing is a fundamental robotic task combining navigation and object manipulation, and has been studied extensively from analytical to learning-based perspectives~\cite{stuber2020let}.
Recent advances in robot learning have further enabled robust visuomotor manipulation policies~\cite{agrawal2016learning,chi2025diffusion}, although primarily in stationary single-robot settings.
Likewise, collaborative box-pushing by multiple mobile robots has long been studied~\cite{mataric1995cooperative}, typically with only a few interacting agents.
In contrast, our coordinated setting enables large-scale interactions involving tens of robots together with end-to-end visuomotor learning.
While vision-based multi-robot coordination has recently emerged~\cite{hu2021vgai,zhang2025learning,mezey2025purely,blumenkamp2024covis}, these studies generally avoid physical object interaction.
Our setting is also related to swarm robotics~\cite{dorigo2021swarm}, studying how large robot teams achieve collective behavior through simple local interactions, while our focus is not on hand-designed interaction rules.

Our work broadly falls within \emph{multi-agent imitation learning (MAIL)}, where policies are learned from demonstrations while incorporating architectural inductive biases for coordinated or collaborative task execution.
When well-structured problem formulations and reliable expert policies are available, MAIL can be implemented via simple behaviour cloning and indeed provides a powerful framework, as demonstrated in domains such as multi-agent pathfinding~\cite{alkazzi2024a,jain2026graph}.
However, deriving such abstractions becomes considerably more difficult in real-world unstructured settings involving object manipulation without access to ground-truth external state information.
Another powerful paradigm is \emph{multi-agent reinforcement learning (MARL)}~\cite{albrecht2024multi}.
Indeed, MARL has demonstrated collaborative box-pushing behaviours for small robot teams~\cite{feng2025learning,zhao2026seqwm}, but relies on nontrivial reward engineering that explicitly encodes desired collective behaviours.
Besides, MARL relies on large amounts of interaction in physics-aware simulation environments, which are costly to develop for visuomotor collectives~\cite{zhang2025learning}.
Consequently, learning visuomotor MAIL policies directly from solo demonstrations becomes an attractive alternative.

Although relatively sparse compared to MARL, several advanced MAIL methods beyond behaviour cloning have been proposed~\cite{le2017coordinated,song2018multi,bhattacharyya2018multi,mai2024inverse,agaskar2025deepfleet}.
In practice, however, the primary bottleneck often lies not in policy learning itself, but in collecting coordinated demonstrations, which constitutes a central challenge in this study.
Closer to our motivation, round-robin behaviour cloning (R2BC)~\cite{mattson2025r2bc} learns heterogeneous multi-agent policies from single-agent demonstrations via sequential on-policy training, albeit only for teams of up to three agents.
Similarly, \cite{ji2025enabling} enables multi-agent hide-and-seek coordination by fine-tuning policies from single-human guidance collected during multi-robot deployments while the remaining robots are controlled by pre-trained coordination-agnostic policies.
Our work follows a similar direction but targets large-scale ego-centric visuomotor policy learning from noisy high-dimensional observations.
Furthermore, these studies rely on top-down cameras and centrally coordinated physical robot deployments.

Given these considerations, our objective is \emph{not} to provide yet another highly engineered coordination mechanism, but rather to provide systematic empirical insights into how current learning techniques perform under large-scale, decentralised visuomotor robot deployments, which has been missing in the literature, stepping towards out-of-the-box multi-robot learning.

\section{Policy Design}
\label{sec:method}
As discussed, coordinated box-pushing is fundamentally solvable by a single robot, while multiple robots primarily improve throughput through concurrent operation.
Our objective is then to acquire robust box-pushing skills---such as finding, pushing, and releasing boxes---while minimising inter-robot interference such as collisions and livelocks.
The challenges include policy architecture design under resource-constrained onboard environments and interaction-aware data collection.
The following sections describe these design choices, while empirical analyses are provided in \cref{sec:exp}.

\subsection{Model Architecture and Training Loss}
\label{sec:method:model}
We adopt a standard encoder--decoder architecture for visuomotor learning (\cref{fig:overview}; middle).
Specifically, MobileNetV2~\cite{sandler2018mobilenetv2} encodes a $96\times72\times3$ RGB image into a latent embedding $\bm{z}_t$ ($36$-dimensional in our setup).
An LSTM~\cite{schmidhuber1997long} then integrates the embedding history $\langle \bm{z}_{t-k}, \ldots, \bm{z}_{t} \rangle$ to predict velocity commands $\bm{y}$, where we use $k{=}5$.
This lightweight design balances model capacity and computational efficiency while retaining history-dependent robot behaviours.
Although transformer-based~\cite{vaswani2017attention} and diffusion-based~\cite{chi2025diffusion} policies were also considered, their inference overhead and latency were incompatible with real-time onboard control on Raspberry Pi-class hardware.

We further enhance the policy capacity using a \emph{mixture-of-experts (MoE)} design~\cite{jacobs1991adaptive}.
Instead of a single decoder, we employ $m$ LSTM expert decoders together with a lightweight router network.
The router applies mean pooling over the embedding history $\langle \bm{z}_{t-k}, \ldots, \bm{z}_{t} \rangle$, followed by a small MLP producing expert weights $w_1,\ldots,w_m$ through softmax normalisation.
The final command is computed as $\sum_{i=1}^{m} w_i \bm{y}_i$, where $\bm{y}_i$ denotes the output of the $i$-th expert.
The policy contains approximately 2.4M parameters for $m{=}2$, of which 2.2M belong to the MobileNetV2 encoder.

The MoE design is motivated by the diversity of behaviours required even in single-robot box-pushing, including box searching, transportation, and recovery from collisions or wall-contact states.
Such multi-expert strategies have been shown to promote behavioural diversity and robust skill acquisition in robotics~\cite{merel2019hierarchical,yang2020multi,hoeller2024anymal}.
While prior studies often pre-train individual skills before learning higher-level routing functions, our approach jointly trains the router and expert decoders end-to-end without manually decomposing behaviours into predefined skills.

To mitigate \emph{expert collapse}~\cite{shazeer2017outrageously,fedus2022switch}, where only a small subset of experts is activated frequently during training, we employ entropy regularisation over the routing weights: $\loss{routing}=\sum_{i=1}^{m} w_i \log w_i$.
Together with the behaviour cloning loss \loss{BC} between the target command $\hat{\bm{y}}_t$ and prediction $\bm{y}_t$, the training objective becomes $\mathcal{L}=\mathcal{L}_{\mathrm{BC}}+\lambda\mathcal{L}_{\mathrm{routing}}$, where $\lambda$ is a balancing hyperparameter.
For \loss{BC}, we use the Huber loss~\cite{huber1964robust} due to its robustness against occasional large control errors.
These design choices are empirically evaluated in \cref{sec:exp:moe}.

\begin{wrapfigure}{r}{0.6\linewidth}
  \vspace{-1.3cm}
  \footnotesize
  \setlength{\tabcolsep}{0.2mm}
  \renewcommand{\arraystretch}{0.85}
  \newcommand{\imgwidth}{0.22\linewidth}
  \centering
  \begin{tabular}{lccccc}
    \toprule
    & \dataset{Single} & \dataset{Static} & \dataset{Dynamic} & \dataset{Recovery}\\
    &
    \includegraphics[width=\imgwidth]{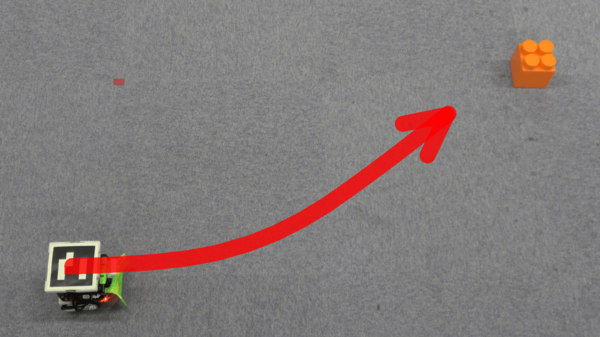} &
    \includegraphics[width=\imgwidth]{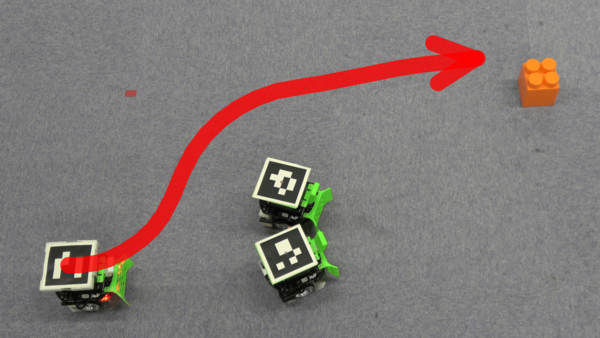} &
    \includegraphics[width=\imgwidth]{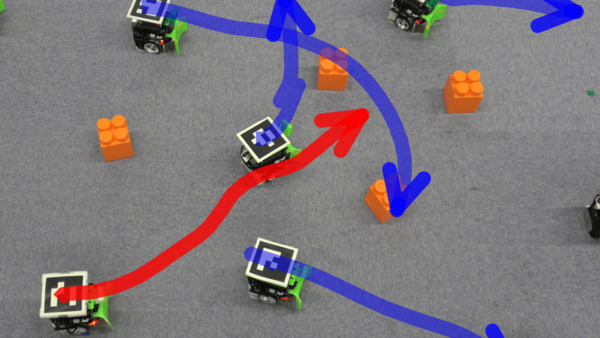} &
    \includegraphics[width=\imgwidth]{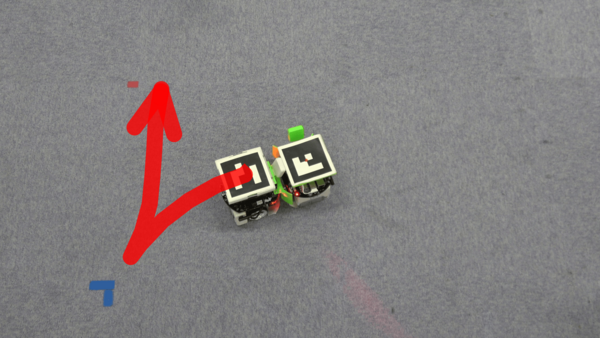}\\
    \midrule
    \model{static} & 6K & 12K & 0 & 2K \\
    \model{dynamic} & 6K & 0 & 12K & 2K \\
    \model{mixed} & 6K & 6K & 6K & 2K \\
    \bottomrule
  \end{tabular}
  \caption{Training dataset breakdowns (\# of observation-action pairs) for each policy.}
  \label{fig:dataset}
\end{wrapfigure}

\subsection{Dataset Creation}
Using a joystick controller, a well-practised operator teleoperates a single robot in the workspace to collect paired observation-action demonstrations.
The challenge here is to determine which dataset types are required and how they should be prepared.

To reduce inter-operator variability, all demonstrations for the same task were collected by a single operator.
Furthermore, for all datasets, the operator directly observes the entire setup during data collection unless otherwise specified.
This choice was intended to facilitate high-quality demonstrations, following common practice in imitation learning for robotic manipulation.
In \cref{sec:exp:priviledged-obs}, we additionally evaluate policies trained on demonstrations collected while the operator relies solely on the live stream from the robot's onboard camera.

Our starting point is the \dataset{Single} dataset, collected using a single robot in a workspace with many randomly distributed boxes.
The operator moves the robot clockwise through the workspace to clean the environment as efficiently as possible.
Each run ends once all boxes have been cleaned up, and this procedure is repeated until the target number of paired observation-action samples is collected (\cref{fig:dataset}).
Training on this dataset alone yields a naive IL policy that is entirely agnostic to the presence of other robots.
When deployed, however, such policies frequently exhibit failure modes such as becoming stuck near walls or trapped by neighbouring robots in multi-robot settings.
To address these cases, rather than relying on dataset aggregation techniques~\cite{ross2011reduction}, we additionally construct a \dataset{Recovery} dataset containing recovery behaviours from such failure situations.
Together, \dataset{Single} and \dataset{Recovery} provide the minimal behavioural primitives for robust single-robot box-pushing.

Training coordinated policies, however, requires data that explicitly reflects the presence of other robots.
The simplest approach is therefore to place \emph{static} robots in the workspace during data collection.
Specifically, we collect solo demonstrations while additional non-controlled robots remain stationary throughout the environment, resulting in the \dataset{Static} dataset.
\emph{As demonstrated later in the experiments, somewhat unexpectedly, even this highly simplified dataset proves sufficient to learn high-performance policies for coordination tasks}.

Motivated by prior studies on learning coordination from solo demonstrations~\cite{mattson2025r2bc,ji2025enabling}, we further construct a dataset termed \dataset{Dynamic}, where observations include other robots actively operating in the workspace.
To obtain this dataset, we adopt a simple bootstrap procedure.
First, we train a stable policy, termed \model{static}, using \dataset{Single}, \dataset{Recovery}, and \dataset{Static}.
We then deploy multiple robots controlled by \model{static} while an operator teleoperates an additional robot in the same environment, resulting in \dataset{Dynamic}.
Compared to \dataset{Static}, this dataset naturally contains richer interaction patterns, including behaviours for avoiding and navigating around moving robots.

In summary, we construct four datasets: \dataset{Single}, \dataset{Static}, \dataset{Dynamic}, and \dataset{Recovery} (\cref{fig:dataset}).
These categories and preparation strategies are designed to remain intuitive yet informative for training MAIL policies, while preserving \emph{transferability} beyond box-pushing.
Indeed, as demonstrated later in \cref{sec:exp:color}, the same strategy can be directly applied to a different coordination task.

\subsection{Auxiliary Future Prediction}
\label{sec:method:future}
Effective multi-agent coordination likely requires agents to anticipate the future behaviours of surrounding robots, especially in communication-free settings.
Such predictive reasoning is closely related to the broader concept of \emph{world models}~\cite{ha2018world}, which have shown strong performance in sequential decision-making~\cite{hafner2025mastering}.
Inspired by this perspective, we incorporate a lightweight world-model-style future prediction objective into policy learning.

Our approach is intentionally simple: predicting future latent embeddings $\langle \bm{z}_{t+1}, \ldots, \bm{z}_{t+l} \rangle$ from past observations $\langle \bm{z}_{t-k}, \ldots, \bm{z}_{t} \rangle$, where $l{=}2$ in the experiments;
an auxiliary LSTM predictor is initialised from the last hidden state of the decoder and recursively rolls out future embeddings in an auto-regressive manner.
We define \loss{world} as the MSE between predicted and ground-truth embeddings.
With this auxiliary objective, the final training loss becomes
$\loss{BC}+\lambda_1\loss{routing}+\lambda_2\loss{world}$,
encouraging the policy to capture environment dynamics and anticipate surrounding agents.

\section{Experiments}
\label{sec:exp}
We systematically investigate learning strategies and further examine whether the same tendencies generalise to a different task.
All results are obtained from real-world robot deployments rather than simulations, with each score averaged over three runs, which we believe provides sufficiently meaningful observations given the substantial deployment effort.
Policy training was conducted on an NVIDIA GeForce RTX 3090.
Most experiments use 20K training samples, corresponding to approximately \SI{1}{\hour} of robot operation.
The \dataset{Static} and \dataset{Dynamic} datasets were collected using 10 robots in the workspace, where one robot was teleoperated while the remaining robots were either stationary or autonomously controlled by \model{static}.
Detailed hyperparameters are provided in the appendix.

\subsection{Dataset and Loss Design}
Our first question is which dataset types and loss designs are essential for coordinated box-pushing, as well as how learned policies behave under larger robot populations (e.g., training with 10 robots but deploying with 20).
To this end, we prepare six policies:
\emph{(i)}~\model{static}, trained mainly on \dataset{Static};
\emph{(ii)}~\model{dynamic}, trained on \dataset{Dynamic};
\emph{(iii)}~\model{mixed}, trained on both datasets;
and \emph{(iv)-(vi)} their future-aware counterparts with \loss{world}.
\Cref{fig:dataset} shows the dataset composition, while \cref{fig:dataset_design} reports throughput and task progress for both the 10-robot in-distribution and 20-robot out-of-distribution settings.

\input{fig/dataset_design}

Our primary observation is that, somewhat unexpectedly, \emph{\model{static} outperforms \model{dynamic}} (10 robots; Welch’s $t$-test, $p=0.027$).
Inspection of deployment videos suggests that \emph{congestion} is a major failure mode:
robots interfere with one another and fail to engage in efficient task execution.
In particular, \model{dynamic} often fails to avoid nearby robots appropriately, causing cascading traffic jams once robots become stuck.
This behavior is also reflected in the average travel distance at the end of the task (\model{static}: \SI[uncertainty-mode = separate]{10.5 +- 2.3}{\meter}, \model{dynamic}: \SI[uncertainty-mode = separate]{8.0 +- 0.5}{\meter})  .

We speculate that this degradation arises because \dataset{Dynamic} is substantially noisier than \dataset{Static}, since surrounding robots operate simultaneously during data collection, resulting in less consistent supervision signals.
In contrast, \dataset{Static} provides more consistent supervision for avoiding others, leading to better interference-minimising policies.
The intermediate performance of \model{mixed} further supports the difficulty of learning from \dataset{Dynamic}.
Note that, albeit trivial, removing \dataset{Static} from the training drops the policy performance; see the appendix.
Meanwhile, \model{static} exhibits a throughput drop in the 20-robot setting, highlighting the difficulty of translating larger robot populations into proportional performance gains.
We conjecture that densely populated situations are underrepresented in the 20K dataset, limiting generalisation to crowded deployments.

Another notable observation is that adding \loss{world} improves performance when \dataset{Dynamic} is included in training.
While these policies still underperform the best \model{static} policy, predicting future states---including surrounding robot behaviours---appears beneficial for coordination.
In contrast, adding \loss{world} to \model{static} degrades performance, likely because future robot behaviours are absent from the underlying demonstrations, rendering the auxiliary prediction task largely noisy.

\subsection{Effect of Operator Observability}
\begin{wrapfigure}{R}{0.48\linewidth}
  \centering
  \includegraphics[width=\linewidth]{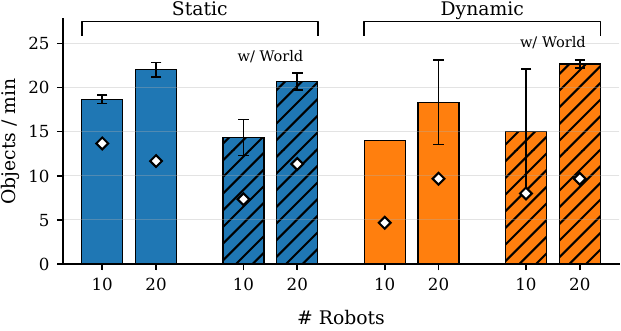}
  \caption{
  Policy performance with data collected via the onboard camera stream. $\diamond$ markers indicate the reference performance obtained with direct operator observation (\cref{fig:dataset_design}).
  }
  \label{fig:privileged_observation}
\end{wrapfigure}

\label{sec:exp:priviledged-obs}

The previous setting relies on privileged observations, where the operator can control the robots based on visual information unavailable to the policy at deployment.
Given this gap between data collection and deployment, we examine how the operator's view during data collection affects policy performance.
Specifically, we collect 20K demonstration samples while restricting the operator to the live stream from the robot's onboard camera, maintaining the dataset proportions in \cref{fig:dataset}.

\Cref{fig:privileged_observation} shows that all evaluated policies substantially outperform their counterparts trained on demonstrations collected with privileged observations.
This indicates that, although demonstrations collected solely from the onboard camera may contain behaviours that are suboptimal from a global perspective, reducing the observation gap between data collection and deployment is crucial for learning effective mobile robot policies.
Nevertheless, the main trends remain unchanged: \model{static} still outperforms \model{dynamic} (10 robots; Welch's $t$-test, $p=0.005$), while \loss{world} continues to improve \model{dynamic}.
These results suggest that effectively leveraging dynamic observations remains challenging even when the operator is restricted to the robot's onboard view.

\subsection{Scaling Training Data}
\label{sec:exp:large-data}
Following the previous results, we next ask: \emph{What happens if we simply increase the amount of data?}
With the priviledged data-collection setting, we repeat the same experiments using three times more training data (60K samples) while keeping the dataset composition unchanged.

\Cref{fig:ablation}a shows the throughput improvement from the 20K setting in \cref{fig:dataset_design}a.
An improvement is achieved for \model{static} in the 20-robot case, mitigating the previously observed dense-population issue.
Meanwhile, scaling the dataset provides only limited gains for \model{dynamic}, suggesting that data scale alone is insufficient for mastering coordination.
Although larger datasets and models may eventually benefit \model{dynamic} through scaling laws, such approaches remain costly in terms of both data collection and onboard computation.
Under these practical constraints, improving data collection, as suggested by the live-stream results, may be a more promising direction for leveraging \dataset{Dynamic}.
Compared to introducing \loss{world}, increasing dataset size generally yields larger performance gains, although the required real-world data collection effort remains substantial.

\input{fig/ablation}

\subsection{Dataset Ablation}
We next investigate the importance of the \dataset{Single} and \dataset{Recovery} datasets in our best \model{static} policy.
These datasets provide solo skills rather than coordination capabilities.
Using 60K samples while keeping the dataset composition unchanged, \cref{fig:ablation}b compares \model{static} with its dataset-ablated variants.

In the 10-robot setting, \dataset{Recovery} appears less critical due to relatively sparse robot interactions.
In contrast, it becomes essential in the 20-robot setting, where robots frequently need to recover from congested situations.
Meanwhile, removing \dataset{Single} still yields functional policies; however, the best 20-robot performance benefits from its inclusion, likely because stronger fundamental box-pushing skills reduce redundant motions and unnecessary interference.

\subsection{Sensitivity to Training Robot Population}
Thus far, we have evaluated policies trained in the 10-robot setting, including out-of-distribution deployment with 20 robots.
Although these settings share the same task structure, denser environments introduce substantially greater coordination difficulty, as commonly observed in coordinated navigation problems~\cite{kato2025congestion,jain2026graph}.
Motivated by this, we further investigate \model{static} under in-distribution 20-robot training and deployment settings.
We prepare two policies trained with 20K samples, where \dataset{Static} is collected using either 10-robot or 20-robot placement configurations.

\Cref{fig:ablation}c shows a natural trade-off:
the policy trained with 10-robot data performs better in the 10-robot setting, whereas the policy trained with 20-robot data performs better in the 20-robot setting.
This indicates clear degradation under out-of-distribution deployment.
We additionally evaluate both policies in an extreme 40-robot scenario, where slight throughput improvements are observed; however, achieving highly efficient coordination remains an open challenge.

\subsection{Importance of Mixture-of-Experts}
\label{sec:exp:moe}
Our backbone policy adopts a mixture-of-experts (MoE) architecture, and we evaluate its effectiveness while varying the number of expert decoders.
\Cref{fig:ablation}d presents this comparison using \model{static} trained on the 20K dataset.
We first confirm that the MoE design is essential for coordinated box-pushing, as using only a single decoder substantially degrades performance.
While the optimal number of experts likely depends on the amount of available training data, we generally find that a small number of experts, such as two, is already sufficient.
Additional analyses of routing regularisation are provided in the appendix.

\subsection{Colour Classification Task}
\label{sec:exp:color}
A recurring observation thus far is that leveraging the \dataset{Dynamic} dataset remains nontrivial.
Indeed, \cref{fig:dataset_design,fig:privileged_observation,fig:ablation} consistently show that \model{static} outperforms both \model{dynamic} and \model{mixed}.
We therefore ask whether this tendency extends beyond the box-pushing task.
To this end, we design a new task termed \emph{colour classification}, where red and yellow boxes are scattered throughout the workspace and must be transported to colour-specific corners.
Task snapshots are shown in \cref{fig:color} (left).
Compared to box-pushing, this task additionally requires colour recognition and transport toward non-unique goal locations, increasing both perception and coordination complexity.
Crucially, the task still constitutes a \emph{coordination} problem, where efficient concurrent execution improves overall throughput.

\input{fig/coloring}

\Cref{fig:color} (right) compares the task completion counts of \model{static} and \model{mixed}, trained using approximately 15K samples collected with 5-robot deployments and a dataset composition similar to \cref{fig:dataset}.
We observe the same trend: \model{mixed} does not outperform \model{static} in either the 3-robot or 6-robot case.
This further supports our earlier finding that, for certain coordination tasks, data containing actively operating robots---at least when generated using \model{static} as a proxy---is not necessarily beneficial for improving communication-free coordination policies.

\section{Discussions}
Focusing on coordinated box-pushing, we investigated learning coordination from solo demonstrations for communication-free visuomotor policies deployable onboard under resource constraints.
A distinctive aspect is its large-scale real-world deployment study.
While detailed empirical analyses have already been presented in \cref{sec:exp}, several broader observations are highlighted below.

First, increasing the number of deployed robots does not automatically translate into proportional performance gains due to congestion effects, as commonly discussed in large-scale multi-robot navigation~\cite{kato2025congestion,jain2026graph}.
Our findings, therefore, suggest that one of the central challenges for visuomotor multi-robot policies is achieving (sub-)linear throughput gains under increasing robot populations.

Achieving this likely requires robots to anticipate others' behaviours and act in ways that maximise overall social welfare.
This naturally raises two questions: \emph{(i)}~how such data should be collected, and \emph{(ii)}~how such behaviours should be learned.
For the latter, our experiments suggest that future latent-state prediction via \loss{world} provides useful supervision.
Meanwhile, for the former, our results suggest that the bottleneck lies not merely in data quantity, but in the quality of interaction distributions:
while considerably easier to collect, static-agent demonstrations provide a surprisingly strong policy, whereas bootstrapped dynamic-agent demonstrations often amplify congestion effects.

We note that our preliminary investigations additionally involved more complicated training procedures, including multi-round bootstrapping where previously trained policies are deployed as proxies akin to R2BC~\cite{mattson2025r2bc}, fine-tuning \model{static} using the \dataset{Dynamic} dataset, task transfer with more diverse data, and combinations thereof.
However, these attempts did not substantially change the overall findings reported throughout this paper.
To conclude, \emph{we believe that leveraging dynamic observations of other robots will ultimately be important for achieving linear performance scaling, which constitutes a key challenge in multi-robot research}.

\section{Limitations}
This study does not target tightly coupled collaborative tasks requiring strongly interdependent multi-robot behaviours.
Instead, our focus is on scalable coordination under realistic onboard resource constraints.
While richer computation and larger models may alter some of the conclusions, such extensions would inevitably introduce non-negligible infrastructural costs.
Likewise, introducing inter-robot communication is not addressed in this study.
Although potentially attractive, it would require substantial additional engineering effort, particularly at the scale considered here.

\section*{Acknowledgments}
The authors would like to thank the anonymous reviewers and the area chair for their insightful comments, and Mizuki Yamaguchi for his assistance with the robot deployment.
This study was partially supported by JST PRESTO (JPMJPR2513).

\bibliography{sty/ref-macro,reference}

\newpage
\appendix
\section*{Appendix}
\section{Experimental Setups}
\paragraph{Mobile Robot}
The robots used in the experiments are based on Raspberry Pi Mouse V3 equipped with Raspberry Pi Camera Module 3~\cite{yamamoto2025100mouse}.
Each robot has a width of approximately \SI{14}{\centi\meter}. 

\paragraph{Box Collection Task}
Robots transport coloured blocks toward a target hole located at the centre of the workspace. 
\Cref{fig:workspace}~(left) shows the workspace layout. 
The experimental area consists of a $\SI{3.5}{\meter} \times \SI{3.5}{\meter}$ indoor space surrounded by white walls. 
At the centre of the workspace, a $\SI{1}{\meter} \times \SI{1}{\meter}$ black carpet region is placed, with a $\SI{0.5}{\meter} \times \SI{0.5}{\meter}$ hole of depth \SI{14.5}{\centi\meter} located at its centre. 
A tower constructed from coloured blocks is positioned at the centre of the hole for visual recognition. 
The tower consists of yellow, blue, and red layers from top to bottom, and is placed such that the distance between the tower and the edge of the hole is $\SI{13}{\centi\meter}$.
The transported objects are 28 orange blocks with dimensions of $\SI{8}{\centi\meter} \times \SI{8}{\centi\meter} \times \SI{10}{\centi\meter}$. 
Task completion is measured using an overhead camera installed above the workspace. 
A transported block is regarded as successfully delivered if it either falls into the hole or if the centre of the block reaches the black carpet region surrounding the hole.

\paragraph{Colour Classification Task}
Robots transport each block to a corner goal with the corresponding colour. 
The experimental area consists of a $\SI{2}{\meter} \times \SI{2}{\meter}$ indoor space surrounded by white walls, with a black carpet covering the floor (\cref{fig:workspace}; right). 
At the four corners of the workspace, red and yellow blocks with dimensions of $\SI{16}{\centi\meter} \times \SI{24}{\centi\meter} \times \SI{10}{\centi\meter}$ are alternately placed. 
These blocks serve as colour-specific corner goals in the task. 
Red and yellow blocks are scattered throughout the workspace. 
The transported objects consist of five red blocks with dimensions of $\SI{8}{\centi\meter} \times \SI{12}{\centi\meter} \times \SI{10}{\centi\meter}$ and five yellow blocks with dimensions of $\SI{8}{\centi\meter} \times\SI{8}{\centi\meter} \times \SI{10}{\centi\meter}$. 
Task completion is measured using a USB camera mounted on a tripod. The number of blocks successfully delivered to the correct corner goals is manually counted every \SI{10}{\second}.

\input{fig/workspace}

\section{Hyperparameters}
\cref{tab:hyperparameters} summarises hyperparameters used for the model architecture and training loss described in \cref{sec:method:model}, as well as the auxiliary future prediction module introduced in \cref{sec:method:future}. 
For the experiments with auxiliary future prediction, the auxiliary world-model loss was enabled after an initial warm-up phase and linearly increased to the final coefficient $\lambda_2$. 
The warm-up schedule depended on the dataset size: for datasets with 20K samples, the world-model loss was enabled from epoch 10 and ramped over 10 epochs, whereas for datasets with 60K samples, it was enabled from epoch 3 and ramped over 10 epochs.
These were adjusted following the pilot studies.

\begin{table}[t]
\centering
\small
\renewcommand{\arraystretch}{0.95}
\caption{Model and training hyperparameters}
\begin{tabular}{lc}
\toprule

\multicolumn{2}{c}{\textbf{Input Configuration}} \\
\midrule
Sequence length & 5 \\
Image resolution & $96 \times 72$ \\

\midrule
\multicolumn{2}{c}{\textbf{Backbone and Encoder}} \\
\midrule
Backbone & MobileNetV2 \\
Encoder output dimension & 36 \\

\midrule
\multicolumn{2}{c}{\textbf{Mixture-of-Experts}} \\
\midrule
Number of experts $m$ & 2 \\
Expert type & LSTM \\
LSTM hidden dimension & 128 \\
LSTM layers & 1 \\
Router MLP hidden dimension & 64 \\

\midrule
\multicolumn{2}{c}{\textbf{Training}} \\
\midrule
Optimizer & AdamW \\
Batch size & 128 \\
Learning rate & $1\times10^{-4}$ \\
Weight decay & $1\times10^{-4}$ \\
Epochs & 30 \\

\midrule
\multicolumn{2}{c}{\textbf{Loss Function}} \\
\midrule
Behaviour cloning loss \loss{BC} & Huber loss \\
Huber $\delta$ & 1.0 \\
$\lambda_1$ (routing loss coefficient) & 0.005 \\
$\lambda_2$ (world loss coefficient) & 0.0005 \\
Future prediction horizon $l$ & 2 \\

\bottomrule
\end{tabular}
\label{tab:hyperparameters}
\end{table}

\section{Effect of Routing Regularisation} 
To evaluate the effect of the routing regularization term \loss{routing} in the loss function, we trained a two-expert MoE model without this term (i.e., using only \loss{BC}). The ablated model was trained on the 20K dataset used for training \model{static}. 
Without \loss{routing}, the routing entropy on the validation dataset did not decrease during training, whereas the model trained with \loss{routing} exhibited decreasing routing entropy. 
We further evaluated whether this difference affected real-world performance by deploying both models to a system consisting of 10 robots. 
The model trained without \loss{routing}exhibited lower task performance compared to the model trained with \loss{routing} (\cref{fig:loss_design}a, MoE models).

\section{Parameter-Matched Comparison between LSTM and MoE Policies}
The results in \cref{sec:exp:moe} support the effectiveness of the MoE architecture. 
However, increasing the number of experts also increases the total number of parameters. 
To control for this effect, we compared the two-expert MoE policy with a parameter-matched single-LSTM policy whose hidden dimension was adjusted (\texttt{hidden}=191) to achieve a comparable parameter count. 
The parameter-matched LSTM policy was trained on the 20K dataset for \model{static} and evaluated on the 10-robot deployment.
Although the parameter-matched LSTM slightly improved performance compared to the original single-LSTM model, it still underperformed relative to the two-expert MoE policy (\cref{fig:loss_design}a). 

\input{fig/appendix_moe2_evaluation}

\section{Effect of Removing Static Data}
To evaluate the contribution of \dataset{Static} data, we trained a policy, denoted as \model{single}, using a dataset consisting only of 15K \dataset{Single} samples and 5K \dataset{Recovery} samples. 
As expected, \cref{fig:loss_design}b shows that \model{single} achieved lower performance than \model{static} in the 10-robot evaluation. 
In contrast, little performance difference was observed between the two policies in the 20-robot evaluation. 
This suggests that, while \dataset{Static} data are beneficial under conditions similar to those represented in the training data, their effect becomes limited in the denser 20-robot environment. 
As discussed in \cref{sec:exp:large-data}, this trend may also depend on the amount of training data available.

\end{document}